\documentclass[final]{siamonline1116}

\usepackage{amsfonts,bm}
\usepackage{graphicx}
\usepackage{epstopdf}
\usepackage{algorithmic}
\ifpdf
  \DeclareGraphicsExtensions{.eps,.pdf,.png,.jpg}
\else
  \DeclareGraphicsExtensions{.eps}
\fi

\numberwithin{theorem}{section}

\newcommand{\TheTitle}{Automated Learning of Boolean Networks} 
\newcommand{\TheAuthors}{J. Sun, A.A.R AlMomani, E. Bollt}

\headers{\TheTitle}{\TheAuthors}

\title{{Data-Driven Learning of Boolean Networks and Functions by Optimal Causation Entropy Principle (BoCSE)}\thanks{Submitted to the editors DATE.
\funding{This work was funded in part by the U.S. Army Research Office grant W911NF-16-1-0081 and the Simons Foundation grant 318812.}}}

\author{
Jie Sun\thanks{Department of Mathematics, Clarkson University, Potsdam, NY 13699, and Clarkson Center for Complex Systems Science, Potsdam, NY 13699}
\and
Abd AlRahman R AlMomani\thanks{Clarkson Center for Complex Systems Science, Potsdam, NY 13699, and Department of Electrical and Computer Engineering, Clarkson University, Potsdam, NY 13699}  
\and
Erik Bollt\thanks{Clarkson Center for Complex Systems Science, Potsdam, NY 13699, and Department of Electrical and Computer Engineering, Clarkson University, Potsdam, NY 13699}   
}

\ifpdf
\hypersetup{
  pdftitle={\TheTitle},
  pdfauthor={\TheAuthors}
}
\fi

\begin{document}

\maketitle

\begin{abstract}
Boolean functions and networks are commonly used in the modeling and analysis of complex biological systems, and this paradigm is highly relevant in other important areas in data science and decision making, such as in the medical field and in the finance industry. In a Boolean model, the truth state of a variable is either 0 or 1 at a given time.  Despite its apparent simplicity, Boolean networks are surprisingly relevant in many areas of application such as in bioinformatics to model gene expressions and interactions.  In the latter case, a gene is either ``on" or ``off" depending on its expression level. Despite the promising utility of Boolean modeling, in most practical applications the Boolean network is not known.
Automated learning of a Boolean network and Boolean functions, from data, is a challenging task due in part to the large number of unknowns (including both the structure of the network and the functions) to be estimated, for which a brute force approach would be exponentially complex.  In this paper we develop a new information theoretic methodology that we show to be significantly more efficient than previous approaches.
Building on the recently developed optimal causation entropy principle (oCSE), that we proved can correctly infer networks distinguishing between direct versus indirect connections, we develop here an efficient algorithm that furthermore infers a Boolean network (including both its structure and function) based on data observed from the evolving states at nodes.   We call this new inference method, Boolean optimal causation entropy (BoCSE), which we will show that our method is both computationally efficient and also resilient to noise.  Furthermore, it allows for selection of a set of features that best explains the process, a statement that can be described as a networked Boolean function reduced order model. We highlight our method to the feature selection in several real-world examples: (1) diagnosis of urinary diseases, (2) Cardiac SPECT diagnosis, (3) informative positions in the game Tic-Tac-Toe, and (4) risk causality analysis of loans in default status. Our proposed method is effective and efficient in all examples. 
\end{abstract}



\section{Introduction} 

In this paper we consider an important problem in data science and complex systems, that is the identification of the hidden structure and dynamics of a complex system from data. Our focus is on binary (Boolean) data, which commonly appears in many application domains. For example, in quantitative biology, Boolean data often comes from gene expression profiles where the observed state of a gene is classified or thresholded to either ``on" (high expression level) or ``off" (low to no expression level). In such an application, it is a central problem to understand the relation between different gene expressions, and how they might impact phenotypes such as occurrence of particular diseases. The interconnections among genes can be thought of as forming a Boolean network, via particular sets of Boolean functions that govern the dynamics of such a network. 
The use of Boolean networks has many applications, such as for the modeling of plant-pollinator dynamics~\cite{campbell2011network}, yeast cell cycles~\cite{li2004yeast,davidich2008boolean}, pharmacology networks~\cite{irurzun2017advanced}, tuberculosis latency~\cite{hegde2012understanding}, regulation of bacteria~\cite{maclean2010boolean}, biochemical networks~\cite{helikar2011boolean}, immune interactions~\cite{thakar2010boolean}, signaling networks~\cite{von2014boolean}, gut microbiome~\cite{steinway2015inference}, drug targeting~\cite{vitali2016network}, drug synergies~\cite{flobak2015discovery}, floral organ determination~\cite{azpeitia2014gene}, gene interactions~\cite{alvarez2015proteins}, and host-pathogen interactions~\cite{raman2010systems}.
In general, the problem of learning Boolean functions and Boolean networks from observational data in a complex system is an important problem to explain the switching relationships in these and many problems in science and engineering.

To date, many methods have been proposed  to tackle the Boolean inference problem~\cite{liang1998reveal,lahdesmaki2003learning,ding2005minimum,marshall2007inference,veliz2012algebraic,berestovsky2013evaluation,liu2016knowledge}. 
Notably, REVEAL (reverse engineering algorithm for inference of genetic network architectures), which was initially developed by Liang, Fuhrman and Somogyi in 1998~\cite{liang1998reveal} has been extremely popular. REVEAL blends ideas from computational causality inference with information theory, and has been successfully applied in many different contexts. However, a main limitation of REVEAL is its combinatorial nature and thus suffers from high computational complexity cost, making it effectively infeasible for larger networks.
The key challenge in Boolean inference are due to two main factors: 
(1)
The system of interest is typically large, containing hundreds, if not thousands and more, components; 
 (2) The amount of data available is generally not sufficient for straightforward reconstruction of the joint probability distribution.
In this paper, we propose that  information flow built upon causation entropy (CSE) for identifying direct versus indirect influences, \cite{sun2014causation,sun2014identifying}, using the optimal causation entropy principle (oCSE) \cite{sun2015causal} is well suited to develop a class of algorithms that furthermore enable computationally efficient and accurate reconstruction of Boolean networks and functions, despite noise and other sampling imperfections. Instead of relying on a combinatorial search, our method iteratively and greedily finds relevant causal nodes and edges and the best Boolean function that utilizes them, and thus is computationally efficient. We validate the effectiveness of our new approach that here  we call, Boolean optimal causation entropy (BoCSE) using data from several real-world examples, including for the diagnosis of urinary diseases, Cardiac SPECT diagnosis, Tic-Tac-Toe, and risk causality analysis of loans in default status

This the paper is organized as follows. In Section 2 we review some basic concepts that define  structure and function of Boolean networks and the problem of learning these from data. In Section 3 we present BoCSE as an information-theoretic approach together with a greedy search algorithm with agglomeration and rejection stages, for learning a Boolean network. In Section 4 we evaluate the proposed method on synthetic data as well as data from real-world examples, including those for automated diagnosis, game playing, and determination of causal factors in loan defaults. Finally, we conclude and discuss future work in Section 5, leaving more details on the basics of information theory in the Appendix.

\section{The Problem of Learning a Boolean Network from Observational Data}
\subsection{Boolean Function, Boolean Table, and Boolean Network}
A function of the form
\begin{equation}\label{eq:bfunc}
f:\mathbb{D}\rightarrow\mathbb{B},~\mbox{where~}\mathbb{B}=\{0,1\},
\end{equation}
is called a Boolean function, where $\mathbb{D}\subset\mathbb{B}^k$ and $k\in\mathbb{N}$ is the arity of the function. For an $k$-ary Boolean function, there are $2^k$ possible input patterns, the output of each is either $0$ or $1$. The number of distinct $k$-ary Boolean functions is $2^{2^k}$, a number that clearly becomes extremely large, extremely quickly, with respect to increasing $k$. Consider for example, $2^{2^3}=256$, $2^{2^5}\approx4.295\times 10^9$, and $2^{2^8}\approx1.158\times10^{77}$ (comparable to the number of atoms in the universe, which is estimated to be between $10^{78}$ and $10^{82}$).  This underlies the practical impossibility of approaching a Boolean function learning problem by brute force exhaustive search.

\begin{figure}[h]
\centering
\includegraphics[width=0.85\textwidth]{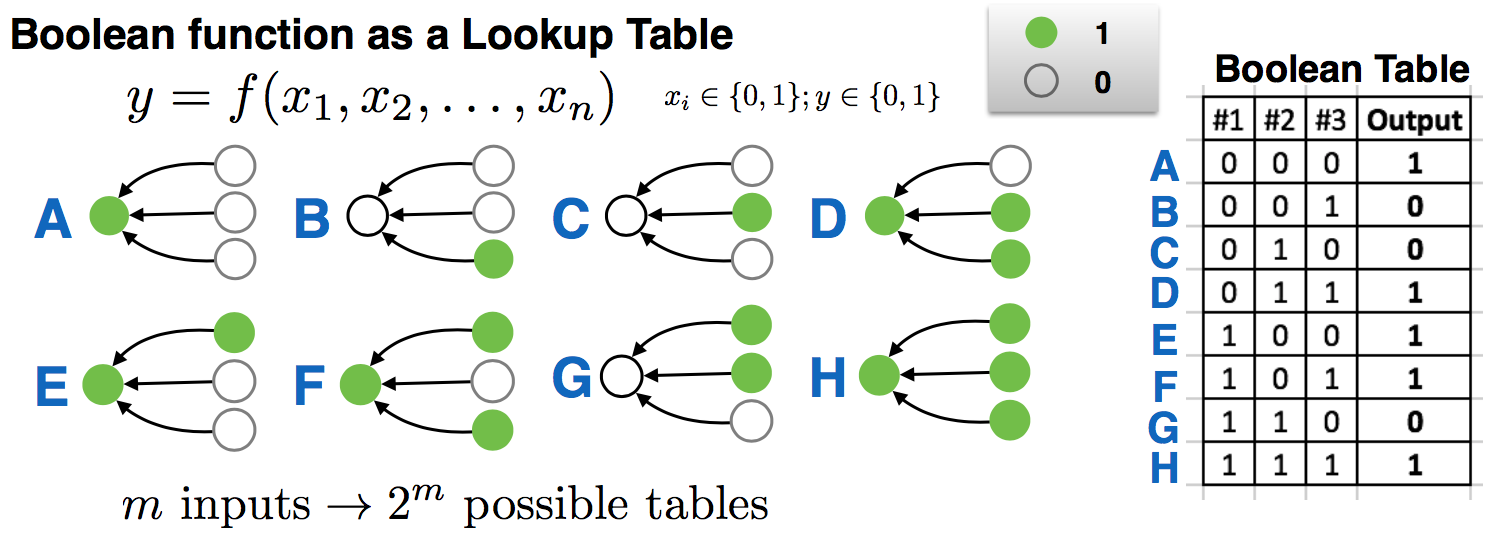}
\caption{A Boolean function can be uniquely identified by a truth table, called a Boolean table. For a $k$-ary Boolean function, the table has $2^k$ rows, and $k+1$ columns. For each row, the first $k$ entries correspond to a particular input binary string (e.g., $(0,1,0)$), and the last entry represents the output of the function.}
\label{fig:btable}
\end{figure}

Each Boolean function can be represented by a truth table, called a {\it Boolean table}. The table identifies, for each input pattern, the output of the function. An example Boolean function $y=f(x_1,x_2,x_3)$ together with its corresponding Boolean table are shown in Fig.~\ref{fig:btable}.

A Boolean function is useful to model a system that has multiple binary inputs and a single binary output. More generally, a system can have multiple outputs, in which case multiple Boolean functions are needed, each capturing the relation between an output variable and the set of input variables. The collection of these individual functions constitute a {\it Boolean network}. Formally, a Boolean network is characterized by a triplet of sets, denoted as $G=(V;E;F)$, where $(V;E)$ represents a graph that encodes the structure of the network: $V(G)=\{1,2,\dots,n\}$ is the set of nodes, and $E(G)\subset V\times V$ is the set of directed edges (possibly including self-loops). The functional rules of the network are encoded in $F(G)=(f_1,f_2,\dots,f_n)$, which is an ordered list of Boolean functions. For each node $i$ in the network, we represent its set of {\it directed neighbors} by $\mathcal{N}_i=\{j:(i,j)\in E\}$ and the degree of node $i$ as the cardinality of $\mathcal{N}_i$, denoted as $k_i=|\mathcal{N}_i|$.
Thus, $f_i:\mathbb{B}^{k_i}\rightarrow\mathbb{B}$ is a $k_i$-ary Boolean function that represents the dependence of the state of node $i$ on the state of its directed neighbors. Note that alternatively the dependence patterns of a Boolean network can also be represented by an adjacency matrix $A=[A_{ij}]_{n\times n}$, where:
\begin{equation}
A_{ij}=
\begin{cases}
1,~\mbox{if $j\in\mathcal{N}_i$;}\\
0,~\mbox{otherwise.}
\end{cases}
\end{equation}
Thus, the adjacency matrix $A$ encodes the structure of a Boolean network, although not the functional rules.

\subsection{Stochastic Boolean Function and Stochastic Boolean Network}
In practice, the states and dynamics of a system are almost always subject to noise. Therefore, it is  important to incorporate randomness and stochasticity into a Boolean network. To do so, we first extend the  Boolean function concept from the deterministic definition to a stochastic generalization, defining a {\it stochastic Boolean function (SBF)} as 
\begin{equation}
	g(x)=f(x)\oplus\xi,
\end{equation}
where $f$ is a (deterministic) Boolean function and $\xi$ is a Bernoulli random variable that controls the level of randomness of the function. In this model, the function contains a deterministic part, given by the Boolean function $f(x)$; the actual output of the function $g(x)$ is given by the output of $f(x)$ subject to a certain probably of being switched.

Following the notion of a stochastic Boolean function, we now define a {\it stochastic Boolean network} as a quadruple of sets, $G=(V;E;F;\bm{q})$, where the triplet of sets $(V;E;F)$ represents a (deterministic) Boolean network, and the vector $\bm{q}=[q_1,\dots,q_n]^\top\in[0,1]^n$ represents the level of noise, each as a random variable each with $q_i$ quantifying the probability of switching the output  state at node $i$, a scalar parameter describing the Bernoulli random variable $\xi_i\sim Bernoulli(q_i) $. 

\subsection{Data from Boolean Functions and Boolean Networks}
We start by discussing several forms of data that commonly appear in application problems. These include: 
(a) Input-output data from a single Boolean function;
(b) Input-output data from a Boolean network, which can be regarded as a generalization of (a);
(c) Time series data from a Boolean network. 
In each one of these scenarios, the data can either be directly represented or rearranged into a set of input-output pairs
\begin{equation}\label{eq:inout}
\{(\bm{x}(t),\bm{y}(t)):t=1,\dots,T\},
\end{equation}
where,
\begin{eqnarray}\label{data1}
\bm{x}(t)&=&[x_1(t),\dots,x_n(t)]^\top\in\mathcal{B}^n=\{0,1\}^n, \mbox{ and, } \nonumber \\
\bm{y}(t)&=&[y_1(t),\dots,y_{\ell}(t)]^\top\in\mathcal{B}^\ell=\{0,1\}^\ell,
\end{eqnarray}
are both vectors of Boolean states. We expand our discussion on this below.

\medskip\noindent
{\bf (a) Input-output data from a single Boolean function.}
For a Boolean function (either deterministic or stochastic), if observations or measurements are made about its inputs and outputs, such data can be represented in the form of~\eqref{eq:inout} where $\bm{y}(t)$ is a scalar (i.e., $\ell=1$).
Here each pair $(\bm{x}(t),\bm{y}(t))$ represents the observed input string of $k$ bits, encoded in $\bm{x}(t)$, and the corresponding output $\bm{y}(t)$. The ordering of the input-output pairs is arbitrary.

\medskip\noindent
{\bf (b) Input-output data from a Boolean network.}
For a (deterministic or stochastic) Boolean network of $n$ nodes, input-output data of the network comes in the form similar to that of a single Boolean function, except that each output itself it no longer a single bit, but instead multiple bits representing the state of all the nodes in the network. Thus, the dimensionality of $\bm{x}(t)$ and $\bm{y}(t)$ are both equal to $n$, that is, $\ell=n$ in the general form of~\eqref{eq:inout}.
The ordering of input-output pairs is arbitrary.

\medskip\noindent
{\bf (c) Time series data from a Boolean network.}
For a time series observed on a Boolean network of $n$ nodes, we can represent such data using a sequence of Boolean vectors $(\bm{x}(t))_{t=0}^{T}$,
where $\bm{x}(t)=[x_1(t),\dots,x_n(t)]^\top\in\mathcal{B}^n$ represents the state of the entire network at time $t$. The pair $(\bm{x}(t-1),\bm{x}(t))$ can be described as an input-output data pair from the underlying Boolean network.
For this matter, time series data from a Boolean network can also be put into the input-output data form~\eqref{eq:inout} with the additional constraint that 
\begin{equation}
\bm{y}(t)=\bm{x}(t+1) \mbox{ for every } t=1,\dots,T-1.
\end{equation}
Here, unlike the case of input-output data as in (a) and (b), the temporal ordering in the time series data is unique and should not be (arbitrarily) changed.

To summarize, in these three commonly encountered scenarios as we discussed above, observational data from a Boolean network can be represented as input-output pairs as in~\eqref{eq:inout}. When the network contains only one node it is really just a Boolean function and thus each $\bm{y}$(t) is a scalar; on the other hand, when the data comes from time series then each $\bm{y}(t)=\bm{x}(t-1)$ and the temporal ordering of the data becomes fixed.

\subsection{The Problem of Learning the Structure and Function of a Boolean Network}
Given Boolean data in the standardized form~\eqref{eq:inout}, we interpret data as samples of a multivariate conditional probability distribution
\begin{eqnarray}\label{eq:pyx}
	p(\bm{y}|\bm{x}) &=& \mbox{Prob}({Y(t)=\bm{y}|X(t)=\bm{x}})\\
	&=& \prod_{i=1}^{k}\mbox{Prob}(Y_i(t)=y_i|X(t)=\bm{x}) = \prod_{i=1}^{k}p(y_i|\bm{x}),
\end{eqnarray}
where $\bm{x}\in\mathcal{B}^k$ and $\bm{y}\in\mathcal{B}^\ell$, and thus \begin{equation}
 p(y_i|\bm{x})=p(y_i|x_1,\dots,x_n). 
 \end{equation}
The problem of reconstructing, or learning the Boolean network then is, can $p(y_i|\bm{x})$ be maximally reduced to a lower dimensional distribution.  That is, does there exist a smallest (sub)set of indices,
\begin{equation}
S_i\subset\{1,\dots,\ell\}, \mbox{ such that }p(y_i|\bm{x})=p(y_i|\bm{x}_{S_i})?
\end{equation}Once we have identified, for each $i$, this set of nodes $S_i$, they together constitute a network, where a directed link $j\rightarrow i$ corresponds to having $j\in S_i$.   Furthermore, to identify such a subset of explaining variables that closely approximates this conditional equality statement represents a simplified or reduced order presentation of the process.

\section{BoCSE for Data-Driven Learning of the Structure and Function of Boolean Networks}
In this section we develop a computational framework to reconstruct both the {\it structure} and {\it function} of a Boolean network from observational data. We start with the reconstruction of a {\it minimally sufficient} Boolean function from input-output data. This method is repeated to find the neighbor set and function for each node, and as a result reconstructs the whole network.

\subsection{Reconstruction of a Minimally Sufficient Boolean function}
Given a set of input-output pairs $\{\bm{x}(t),y(t)\}$, (here $y(t)$ is a single bit), we want to find a minimal Boolean function that is sufficient in representing the data. To quantify the complexity of the Boolean function, we state the following  information-theoretic criterion
\begin{equation}
\begin{cases}
\min_{K\subset[n]}|K|,\\
\mbox{s.t.}~I(X^{(K)};Y)=\max_{K\in[n]}I(X^{(K)};Y)
\end{cases}
\end{equation}
Here,
\begin{eqnarray}
[n&=&\{1,2,\dots,n\}, \nonumber \\
K&=&\{k_1,\dots,k_\ell\} \mbox{ is a subset of }[n], \nonumber \\
Y&=&[y(1),\dots,y(t)]^\top, \mbox{ and }\nonumber \\
X^{(K)}&=&[X^{(K)}]_{T\times\ell} \mbox{ where }
[X^{(K)}]_{tj} = X(t)_{k_j}.
\end{eqnarray}
The symbol $I$ denotes mutual information, that is, $I(X^{(K)};Y)$ is the mutual information between $X^{(K)}$ and $Y$.

At a glance, solving this combinatorial problem seems to be computationally complex. However, in our previous work~\cite{sun2015causal} we developed an oCSE algorithm that can find $K$ efficiently, and we proved in \cite{sun2015causal} that it correctly infers the underlying network as it is able to distinguish direct versus indirect connections correctly.  Here we will further develop the concept to also learn the associated Boolean functions on the networks, that here we call  BoCSE. Although various extensions of the oCSE algorithm are possible, some may even yield better results in certain scenarios.  We focus here on the most basic version of our otherwise greedy search algorithm that consists of only two stages, a forward selection stage and a backward elimination stage.  
\begin{itemize}
\item Forward selection. We initialize the solution set $K_f=\emptyset$, and, in each iteration, we choose an element $k$ that satisfies the following conditions
\begin{equation}\label{eq:forward}
\begin{cases}
\max_{j}I(X_j;Y|X^{(K_f)})>0,\\
k = \arg\max_{j}I(X_j;Y|X^{(K_f)}).
\end{cases}
\end{equation}
If such a $k$ exists, then we append it to the set $K_f$ and proceed to the next iteration; otherwise, when no such $k$ exists, the forward selection is terminated.
\item Backward elimination. Start with $K_b=K_f$, in each step of backward elimination, we select an element $k$ that satisfy the following
\begin{equation}\label{eq:backward}
k = \arg\min_{j\in K_b}I(X_j;Y|X^{(K_b/\{j\})}).
\end{equation}
Such $k$ always exists since $K_b$ is a finite set. Then, if
\begin{equation}
	I(X_k;Y|X^{(K_b/\{k\})})=0, 
\end{equation}
we remove $k$ from $K_b$ and repeat; otherwise, the algorithm terminates.
\end{itemize}
The result of the algorithm is a set $K_b=\{k_1,\dots,k_\ell\}$, which is an estimate of the index set of the minimal Boolean function that fits data. Finally, given such a set $K_b$, we construct the corresponding Boolean function by estimating the best output ($0$ or $1$) for each unique input pattern available from the data. Symbolically, for each $\bm{x}_0\in\mathbb{B}^\ell$, we define the set
\begin{equation}
\mathcal{T}_{K_b}(\bm{x}_0)=\{t:\bm{x}^{(K_b)}(t)=\bm{x}_0\},
\end{equation}
and define
\begin{equation}
g(\bm{x}_0)=\frac{\sum_{t\in\mathcal{T}_{K_b}(\bm{x}_0)}y(t)}{|\mathcal{T}_{K_b}(\bm{x}_0)|}
\in[0,1].
\end{equation}
Then, we obtain $f:\mathbb{B}^\ell\rightarrow\mathbb{B}$ using the tabular form, by defining
\begin{equation}
f(\bm{x}_0)=\lceil g(\bm{x}_0)\rceil\in\{0,1\}.
\end{equation}
If $\mathcal{T}_{K_b}(\bm{x}_0)=\emptyset$ for some $\bm{x}_0$, it means that particular input pattern is never observed in the data. Then, in the absence of additional information, the value of $f$ for such input cannot be optimally determined (the choice of either $f(\bm{x}_0)=0$ or $f(\bm{x}_0)=1$ makes no difference in ``fitting" the data).

\subsection{Estimation of Conditional Mutual Information and Tests of Significance}
The proposed BoCSE learning approach requires estimating various forms of mutual information and conditional mutual information (see Appendix for their definition) from data.
In practice (that is, when entropies need to be estimated from data), a threshold (either $\varepsilon$ or $\eta$) needs to be determined in each step of either the forward or backward stage of the algorithm. The key is to decide, from data, whether an estimated conditional mutual information of the form $\hat{I}=I(X;Y|Z)$ should be regarded as zero, with confidence (as opposed to positive). In particular, we need to consider
\begin{equation}
\begin{cases}
H_0\mbox{(null hypothesis):~} \hat{I}=I(X;Y|Z)=0,\\
H_1\mbox{(alternative hypothesis): ~} \hat{I}=I(X;Y|Z)>0.
\end{cases}
\end{equation}
To decide whether or not to reject $H_0$ (here equivalent as accepting $H_1$), we construct shuffled data by permuting the time ordering of the components in $X$. To be specific, suppose that 
\begin{equation}
\sigma:\{1,\dots,T\}\rightarrow\{1,\dots,T\}
\end{equation}
is a random permutation function, from which we compute $I(X^\sigma;Y|X_{\hat{S}})$ where $X^{\sigma}$ represents the shuffled time series $\{x_{\sigma(1)},x_{\sigma(2)},\dots,x_{\sigma(T)}\}$. By sampling $\sigma$ uniformly, we then obtain a cdf 
\begin{equation}
F(x)=P(I(X^\sigma;Y|X)\leq x).
\end{equation}
From this cdf, we can then estimate the $p$-value under $H_0$ to be $1-F(\hat{I})$, from which we can determine the threshold. For a given $\alpha$-level (e.g., $\alpha=0.01$), the corresponding threshold can then be decided as
\begin{equation}
\begin{cases}
\varepsilon=F^{-1}(1-\alpha),&\mbox{for forward selection};\\
\eta=F^{-1}(1-\alpha),&\mbox{for backward elimination}.
\end{cases}
\end{equation}
Throughout this paper, we set the same $\alpha=0.05$ for both the forward and backward stage of the algorithm (unless otherwise noted), and obtain the cdf $F(x)$ by  uniformly sampling by selecting $1000$ independent random permutation functions $\sigma$.

\section{Examples of Applications}
In this section we now present examples of applications of BoCSE, the proposed Boolean learning method.

\subsection{Benchmark on Random Boolean Networks}
\begin{figure}[htbp]
\centering
\includegraphics[width=0.47\textwidth]{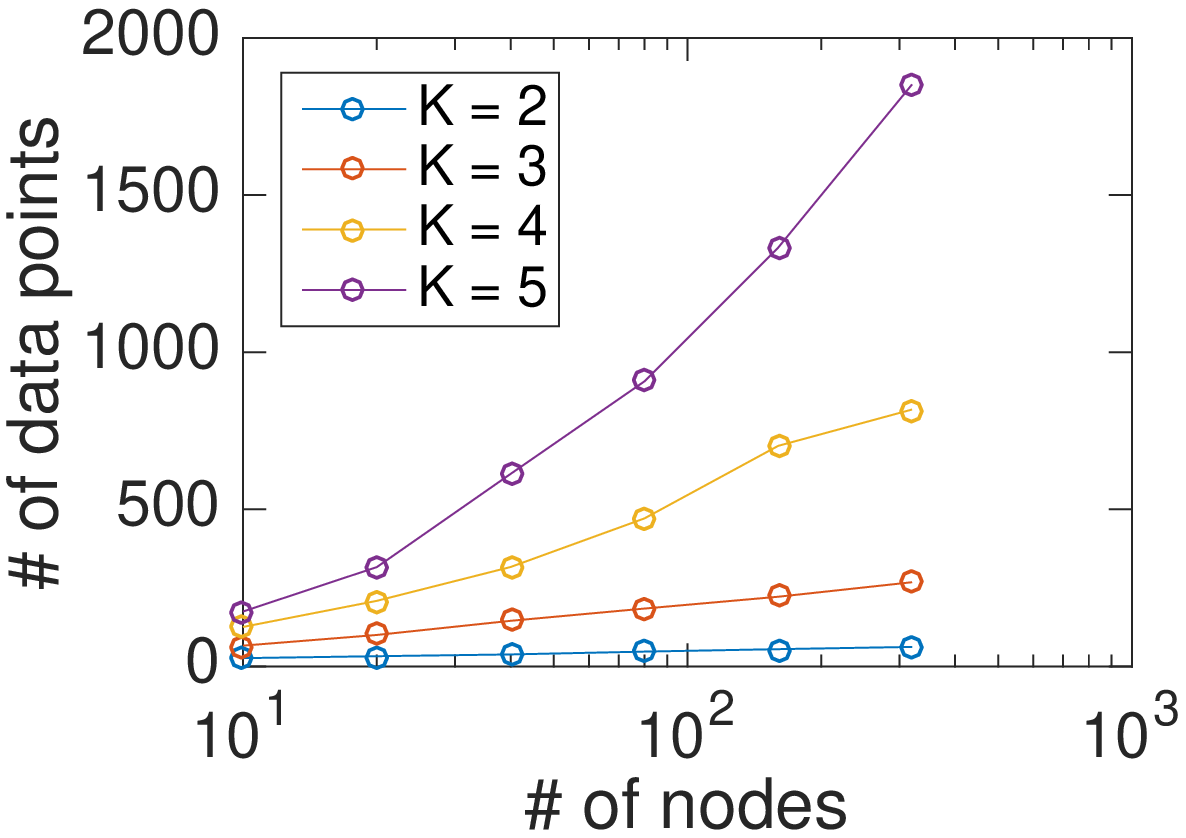}
\includegraphics[width=0.45\textwidth]{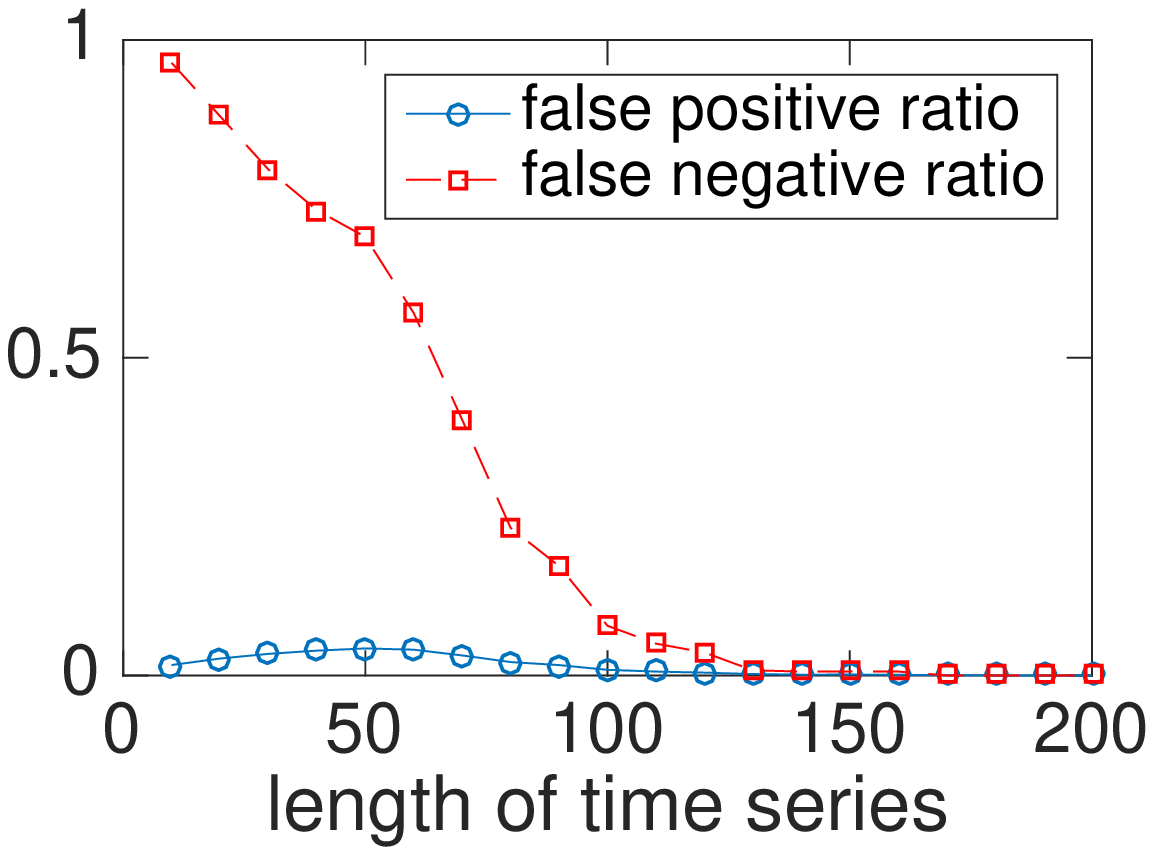}
\caption{(Left) Number of data points required for learning random Boolean networks with no error.  (Right) Error ratios as a result of applying the proposed method BoCSE for learning random Boolean networks of fixed size $n=50$ and degree $K=3$. For both panels, each point on the plotted curve is the result of an average over 50 random realizations.}
\label{fig:random}
\end{figure}

We first evaluate BoCSE for learning randomly generated Boolean networks. These networks are generated with two parameters, $n$ is the number of nodes, and $K$ is the in-degree of each node in the network (for example, $K=3$ means that each node $i$ receives three inputs from other nodes, randomly chosen). The Boolean function associated with each node $i$ is constructed by assigning randomly an output of either $0$ or $1$ to each input pattern, with the equal probability. Figure~\ref{fig:random} shows that, the number of data points needed for correctly learning the entire Boolean network scales sublinearly as the size of the network (left panel). Although the scaling becomes worse as $K$ increases, it is still within practical reach for networks of several hundred of nodes. In the right panel of Fig.~\ref{fig:random}, we show the error of learning for networks of fixed size $n=50$. As the length of time series increases (more data points), both false positive and false negative ratios decrease toward zero, confirming the validity and convergence of the method. 

\subsection{Automated Diagnosis of Urinary Diseases}
As an application to aid the automation of medical diagnosis, we consider a dataset that documents the symptoms and diagnosis outcomes of 120 patients. The data is an extended version of the table used in Ref.~\cite{czerniak2003application}, and is available at the UCI Machine Learning database, via the following link under the name ``Acute Inflammations":~\url{https://archive.ics.uci.edu/ml/datasets/Acute+Inflammations}.

In this extended dataset, there are descriptions from  a total of $120$ patients, each with $6$ attributes and $2$ decision variables. The attributes are: (1) temperature, (2) nausea, (3) lumbar pain, (4) urine pushing, (5) micturition pains, (6) burning of urethra, itch, swelling of urethra outlet. Other than temperature, which takes value in the range of $35$-$42^\circ \mbox{C}$), all the other 5 attributes are recorded as a Boolean value, either ``1'' (symptom exists) or ``0''. In our analysis, we threshold the temperature data into binary values by simple thresholding: temperature equal or above $38^\circ \mbox{C}$ are converted into ``1'' (fever) and those below are converted into ``0'' (no fever).
The two decision (outcome) Boolean variables are
\smallskip
\begin{enumerate}
\item (acute) inflammation of urinary bladder,
\item nephritis of renal pelvis origin.
\end{enumerate}
\smallskip
In Table~\ref{table:urinary} we summarize the description of the attributes and decision variables.

\begin{table}[h]
\centering
\begin{tabular}{ | l | p{4.7cm} |}
\hline    
attributes & description \\ \hline
$X_1$ & fever \\
$X_2$ & nausea\\
$X_3$ & lumbar pain \\
$X_4$ & urine pushing \\
$X_5$ & micturition pains \\
$X_6$ & burning of urethra, itch, swelling of urethra outlet\\\hline
\end{tabular}
\hspace{0.25in}
\begin{tabular}{ | l | p{5cm} |}
\hline    
outcome & description \\ \hline
$Y_1$ & (acute) inflammation of urinary bladder\\
$Y_2$ & nephritis of renal pelvis origin\\
\hline
\end{tabular}
\vspace{0.05in}
\caption[]{Attributes and outcome variables for the urinary disease data. Each variable is Boolean and takes value $1$ or $0$ representing the presence or absence of a particular attribute/outcome.}
\label{table:urinary}
\end{table}

We apply the BoCSE learning method separately to the two outcome variables. For each outcome variable, we treat each patient's attributes as one input Boolean string and the corresponding recorded outcome as a single output.

For the first outcome variable $Y_1$, that is the inflammation of urinary bladder, we found that the relevant attributes are (in terms of decreasing order of importance): (4) urine pushing, (5) micturition pains, and (6) burning of urethra, itch, swelling of urethra outlet. The inferred Boolean function for the relation between these attributes and the outcome are shown in the left part of Table~\ref{table:urinary2}, and is found to accurately describe every individual data record. Interestingly, for the other outcome $Y_2$, the relevant attributes become $X_1$ and $X_3$ (in the order of decreasing importance), and the inferred Boolean function as shown in the right table of Table~\ref{table:urinary2}, can be written using a simple ``and'' gate: $Y_2=X_1 \wedge X_2$, which implies that the diagonals of nephritis of renal pelvis origin can be based on having both symptoms: fever and lumbar pain. Yet again, this relation is consistent with every single patient's record in the dataset.

\begin{table}[h]
\centering
\begin{tabular}{ | c | c | c || c || c |}
\hline    
$X_4$ & $X_5$ & $X_6$ &  $Y_1$ & Occurrence \\ \hline
0 & 0 & 0 & 0 & $25.00\%$\\
0 & 0 & 1 & N/A & $0\%$\\
0 & 1 & 0 & 0 & $8.33\%$\\
0 & 1 & 1 & N/A & $0\%$\\
1 & 0 & 0 & 1 & $8.33\%$\\
1 & 0 & 1 & 0 & $17.50\%$\\
1 & 1 & 0 & 1 & $16.67\%$\\
1 & 1 & 1 & 1 & $24.17\%$\\
\hline
\end{tabular}
\hspace{0.25in}
\begin{tabular}{ | c | c || c || c|}
\hline    
$X_1$ & $X_3$ & $Y_2$  & Occurrence \\ \hline
0 & 0 & 0 & $33.33\%$\\
0 & 1 & 0 & $16.67\%$\\
1 & 0 & 0 & $8.33\%$\\
1 & 1 & 1 & $41.67\%$\\
\hline
\end{tabular}
\vspace{0.05in}
\caption[]{Inferred Boolean relations by BoCSE for the two outcome variables: $Y_1$ (left table) and $Y_2$ (right table). Each entry in the ``occurrence'' column shows the fraction of observed attribute data: $(X_4,X_5,X_6)$ for the left table and $(X_1,X_3)$ for the right table. For each attribute pattern, the ``predicted'' value of outcome is shown in the $Y$ column, where ``N/A'' refers to cases where no such input pattern is ever seen in the empirical data.}
\label{table:urinary2}
\end{table}

\begin{figure}[h]
\centering
\includegraphics[width=0.8\textwidth]{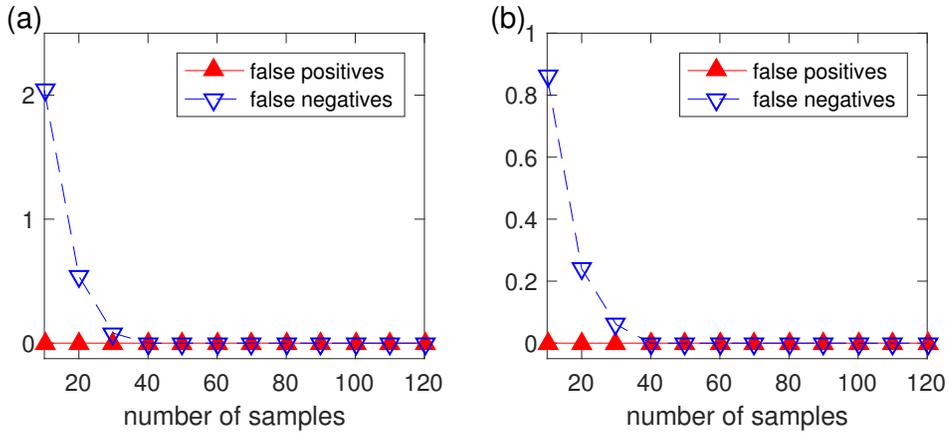}
\caption{Boolean inference error as a function of sample size for the urinary disease example. Here the ``true'' set of relevant attributes are taken to be the ones inferred using the full data (120 samples). (a) Inference error as a function of sample size (from 10 to 120) for the first outcome variable, $Y_1$, where error is quantified by false positives and false negatives. The number of true attributes is 3 in this case. (b) Similar to (a), but for $Y_2$. The number of true attributes is 2 in this case. In each plot, each data point is an average over $50$ independent random down-sampling of the full dataset.}
\label{fig:urinary}
\end{figure}

Next, using the inferred attributes from the entire dataset (120 samples), we explore the dependence of the accuracy of our Boolean inference on the sample size. We do this by randomly selecting a subset of the samples, and use such ``down-sampled'' data instead of the full dataset for Boolean inference. For each sample size, we repeat such inference $50$ times and compute the average number of false positives (attributes inferred using the down-sampled data that are not present using the full-size data) and false negatives (attributed to inferrence using the full  data set which  now appears using the down-sampled data). The results are shown in Fig.~\ref{fig:urinary}. Interestingly, for this particular example our method never seems to produce false positives, and the number of false negatives decrease rapidly to zero as more samples are used in Boolean inference, which suggests effectiveness of the method in automated diagnosis systems via relatively small sample sizes.

\subsection{Automated Cardiac SPECT Diagnosis}
In this example, we test our Boolean learning method on an existing dataset that aims at automated image-based cardiac diagnosis. The dataset is derived from a set of images obtained by 
cardiac Single Proton Emission Computed Tomography (SPECT)~\cite{kurgan2001knowledge}. In particular, there is a total of $T=267$ patients, each of whom is classified as either normal ($y_t=1$) or abnormal ($y_t=0$). The data is divided into a training set which contains $T_1=80$ patients and a test (validation) set of $T_2=187$ patients. For each patient's image set, a total of $n=22$ binary feature patterns were created, defining $x_i(t)=1$ if the $i$-th feature is present in the SPECT images of the $t$-th patient, and $x_i(t)=0$ otherwise, for $i=1,\dots,22$ and $t=1,\dots,267$. Finally, this post-processed Boolean dataset is further divided into a training set which contains $87$ out of the $267$ patients' features and diagnosis, and a validation set which contains such information for the remaining $180$ patients.

\begin{figure}[htbp]
\centering
\includegraphics[width=0.7\textwidth]{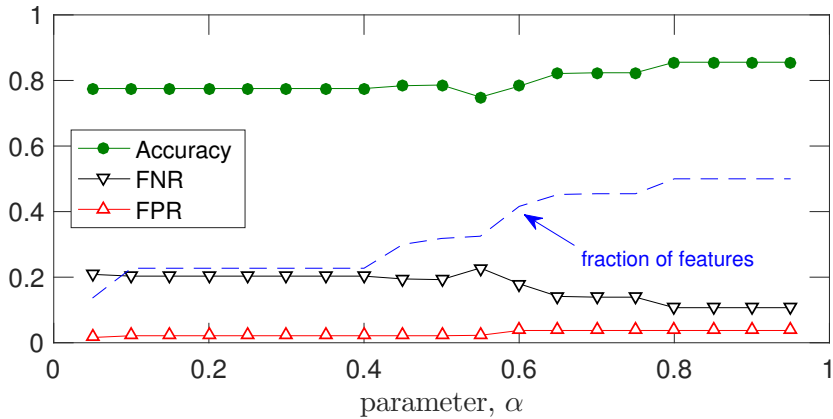}
\caption{Automated diagnosis of heart disease using $22$ Boolean attributes derived from cardiac SPECT. Here we explore how the diagnosis accuracy changes as the parameter $\alpha$ in our Boolean inference method is varied. In particular, we apply BoCSE to the training data (80 patients) and validate the resulting Boolean functions on the validation set (187 patients). We compute the accuracy of diagnosis as the overall percentage of correct diagnosis in the validation set, shown in the figure. In addition, we also compute and plot, for each $\alpha$, false positive ration (FPR) and false negative ratio (FNR), together with the effective number of Boolean variables inferred by our method (dashed curve).}
\label{fig:spectheart}
\end{figure}

Focusing on this post-processed Boolean data, we are interested to see if our automated Boolean inference method is able to learn the decision rules, that is, to diagnose a patient based on a reduced set of Boolean features out of the $22$ features. In this sense, our methodology can be understood as useful for reduced order modelling (ROM) in the realm of complex Boolean function inference problems.  Said  another way, this method describes a way to simplify decision making problems by focusing on the most relevant factors that are those that lead to important outcomes.  As shown in Fig.~\ref{fig:spectheart}, our method is able to learn a Boolean function that achieves near $80\%$ of decision accuracy on the validation data across a wide range of parameters. The our achieved accuracy, generally using only a subset of the full set of $22$ Boolean features, and without any fine-tuning of parameters or further optimization, is already comparable to the best known result on such datasets~\cite{cios2002hybrid}.

\subsection{Tic-Tac-Toe}
\begin{figure}[htbp]
\centering
\includegraphics[width=0.85\textwidth]{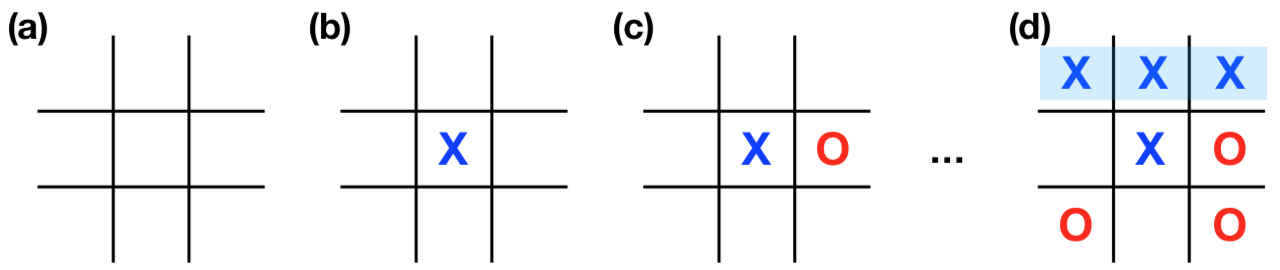}
\caption{Tic-Tac-Toe game. (a) Start of the game, where the board is made up of a 3-by-3 grid. (b-d) Example of a sequence of moves made by two players, where player ``X'' plays first, and eventually won the game by filling in an entire horizontal row.}
\label{fig:tictactoe1}
\end{figure}

The Tic-Tac-Toe is a classical two-player board game, which is also often played on pencil-and-paper. The ``board" is a $3$-by-$3$ grid with a total of $9$ slots, as illustrated in Fig.~\ref{fig:tictactoe1}(a). At the beginning of the game, the board is empty. Then, the two players take turn to mark any empty ``slot" in each turn---typically one uses ``X" the other uses ``O". The player that is the first to have marked three consecutive horizontal, or vertical, or diagonals slots, wins the game. For instance, Fig.~\ref{fig:tictactoe1}(b-d) is an example of the sequence of marks made by the players, where the first player (player ``X'') eventually wins by having marked an entire row (in this case, the top row). In general, if both players do their best at every move, the outcome would be a draw.

Our interest here is not (re)analysis strategies of this relatively simple game. For those who are interested, variants of the game actually has a connection to Ramsey theory, \cite{golomb2002hypercube,patashnik1980qubic}. Here, we are intersted in testing our Boolean learning algorithm to see if it provides any useful information. To this end, we collected the complete set of possible board configurations at the end of a tic-tac-toe game, via the following link under the name ``Tic-Tac-Toe Endgame Data Set ":~\url{https://archive.ics.uci.edu/ml/datasets/Tic-Tac-Toe+Endgame}. There is a total of 958 instances. For the $t$-th instance, we use $\bm{x}(t)=[x_1(t),\dots,x_9(t)]^\top$ to present the state of the $i$-th slot, ordered as follows: upper-left, upper-middle, upper-right, middle-left, center, middle-right, lower-left, lower-middle, and lower-right. Each $x_i(t)$ can either be $1$ (if marked by ``X''), $-1$ (if marked by (``O''), or $0$ (if empty). Corresponding to each instance $t$ is the final outcome, which we denote as $y(t)$, which either equals $1$ if player ``X'' wins or $0$ if ``X'' does not win. Interpreting $\bm(t)$ and $y(t)$ as samples of random variables $\bm{X}$ and $\bm{Y}$, we can ask the question of which slots in the board, statistically, are more relevant (or predicative) for the first player to win the game.

Applying our Boolean learning algorithm, we found a list of most important slots, ordered in decreasing value of (added) relevance: $i_1=5$ (the center), $i_2=1$ (upper-left), $i_3=9$ (lower-right), $i_4=3$ (upper-right), $i_5=7$ (lower-left), $i_6=8$ (lower-middle), and $i_7=2$ (lower-right). To quantify the relative importance of each attribute, we compute the conditional entropy $H(Y|X_{i_1\dots i_k})$ for $k=0,\dots,7$, where $H(Y|X_{i_0})$ is used to represent $H(Y)$. This shows that, as the number of attributes increases, uncertainty decreases monotonically and reaches $0$ (complete predictability) when 7 attributes are used.

\begin{figure}[htbp]
\centering
\includegraphics[width=0.9\textwidth]{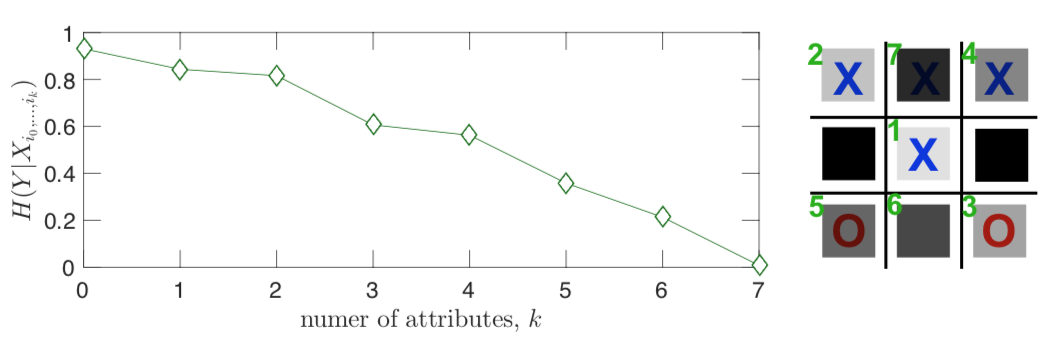}
\caption{Decrease of uncertainty in ``predicting" the outcome of a Tic-Tac-Toe game using partial observations of the board in the final configuration. Here uncertainty is measured by the conditional entropy $H(Y|X_{i_1\dots i_k})$, and the indices $i_k$ are obtained using our Boolean inference algorithm: $i_1=5$ (the center), $i_2=1$ (upper-left), $i_3=9$ (lower-right), $i_4=3$ (upper-right), $i_5=7$ (lower-left), $i_6=8$ (lower-middle), and $i_7=2$ (lower-right).}
\label{fig:tictactoe2}
\end{figure}

\subsection{Risk Causality Analysis of Loans in Default Status}
Loan default prediction is an essential problem for the banks and insurance companies to fiscally responsibly approve loans. However, in many cases, the borrowers fail to pay the loan as agreed, called loan default, which motivates the risk analysis problem in the banking industry, to identify those parameters that identify one borrower as trustworthy, and another borrower as representing a high risk.

We consider the open dataset from LendingClub (American peer-to-peer lending company), which can be downloaded from \href{https://www.lendingclub.com/statistics/additional-statistics?}{LendingClub} website. We considered the dataset for the year 2019 (four quarters). The dataset contains more than 500,000 entries (data points, sample size). However, we only considered the long term (the final) status of the loans.  Therefore,  we excluded all the loans with the status \textit{``Current''} as an outcome, to have a sample size of 62,460 for our analysis. That is, all can be classified to and outcome \textit{``Paid in full''}, or \textit{``Default''} status. We should emphasize here that we only considered the parameters as Boolean in nature, which limits the considered  to those 10 parameters that we investigate as to their influence on the outcomes. 

This example gives a causality driven description of those parameters that \textit{combined=}, can represent a high risk that the borrower will not be able to pay his loan in full.  This causality inference occurs within the 
Boolean framework for parameters concerning the loan long term status.
Table~\ref{table:LoanData} shows the attributes and their description. In loan issuing risk analysis, the amount of the requested loan and the annual income of the borrower are important variables to consider, and they are both numeric variables. We introduce here the combined attribute, loan to income ratio, which combines both variables in the form of a Boolean variable. Our dataset has a sample size of 62,460, and the loan to income ratio range from 0.0001 to 36000. So, we considered the median value $\mu \approx 0.2$ to be our threshold step, such that $X_9 = 0$, indicate that the loan to income ratio of the loan request is less than $0.2$, and it is within the lowest 50\% of all the requested loans over the period (which is in our case, one year).

We then apply BoCSE, the proposed Boolean factors learning approach to the outcome variable $Y$, where we found that the relevant attributes are,  in decreasing order of importance:
\smallskip
\begin{itemize}
    \item $X_9$, Loan to income ratio.
    \item $X_{10}$, Loan terms.
    \item $X_3$, Verification of the reported income.
\end{itemize}
\smallskip

We expect that the probability of having the loan fully paid ($Y = 1$) will be larger than a default ($Y=0$) in this example,  for all the observed combinations of the relevant attributes. However, the challenge here is to find the combination of attributes that, together represent a high risk if approving the loan. For example, if for some attribute combinations (binary string) $X$, the probability $Pr(Y=1) = 0.8$, and $Pr(Y=0) = 0.2$, we may not be satisfied by saying that the expected outcome that the borrower will pay the loan in full because that $Pr(Y=1) > Pr(Y=0)$. Our focus here will be that there is a risk that the borrower will not pay the loan with probability $Pr(Y=0) = 0.2$, which represents a high risk.

In Table~\ref{table:loanResults}, we can see the inferred Boolean function   relationship  between  these  attributes  and  the resulting  outcomes. From application of our automated Boolean function learning method, result shown in  Table~\ref{table:loanResults} we summarize these interesting summary observations:
\begin{itemize}
    \item The first four rows represent patterns where $X_3 = 0$, that describe loans in which the borrower's income was not verified. We see that this feature coincides with a significant increase in the probability that the loan will not be paid in full. The lowest value in this group of patterns is $(X_3, X_9, X_{10}) = (0, 0, 1)$, which represents an unverified income, low loan to income ratio, and 60 months loan term, and the joint probability is then $1.9\%$. We conclude that low loan to income ratio combined with long term loan (which implies low monthly payment) reduces the risk of loan default.
    
    \item For the same pattern, but with a 36 months loan term, $(X_3, X_9, X_{10}) = (0, 0, 0)$, we see a significant increase in risk from $1.9\%$ to $9.6\%$. For the pattern $(X_3, X_9) = (0, 1)$,the risk increases with the 36-month term loan, from $6.3\%$ to $8.1\%$. This indicates that higher monthly payments indicate a higher risk.  However, the effect of the loan term becomes neutral, meaning no effect in terms of observing only the verified income patterns $(X_3 = 1)$. For these two patterns, where we have a verified income,  the same risk conclusions follow for both the 36 and 60 months terms loans.
    
    \item Comparing the above two points, we conclude that if the borrower's reported income is verified, there is no difference in the risk between different loan terms. However, if it is not verified, then going with the 60 months term loans can profoundly reduce the risk, regardless of the loan to income ratio.
    
    \item We see that the lowest risk, or the most trustworthy borrowers, are the ones with the combination $(X_3, X_9, X_{10}) = (1, 0, 1)$, which represent a verified income, low loan to income ratio, and 60 months term loan reflecting low monthly payments. The risk, in this case, is about $0.4\%$. Unfortunately, however, such a pattern occurs infrequently, representing fewer  than $1\%$ of the borrower customers.
    
    \item On the other hand, we see that a significant high risk associates with the combination $(X_3, X_9, X_{10}) = (0, 0, 0)$, which represents unverified income, low LTI ratio, and 36 months term loan (high monthly payments). This is particularly interesting  since a low LTI may on its own may suggest a safer primary indication because it implies a high income, low loan value, or both. However, we see that even with high income, or low requested loan amount, the combination of unverified income together with large monthly payment, $(X_3, X_{10}) = (0, 0)$, has the highest risk as compared to  all other combinations in the table, $9.6\%$ and $8.1\%$.
\end{itemize}

\begin{table}[h]
\centering
\begin{tabular}{ | l | p{12cm} |}
\hline    
attributes & description \\ \hline
$X_1$ & Home ownership. (1: Homeowner, 0: Not a homeowner) \\ \hline
$X_2$ & Delinquency in the past two years. (1: Delinquency occurred, 0: No Delinquency)\\ \hline
$X_3$ & Verification of the reported income. (1: Income verified, 0: Income not verified)\\ \hline
$X_4$ & History of public records. (1: There is a public record, 0: No public records.)\\ \hline
$X_5$ & Application type. (1: Individual, 0: With co-borrower)\\ \hline
$X_6$ & 120 days past due. (1: Have account that ever past due more than 120 days, 0: Never past due more than 120 days.)\\ \hline 
$X_7$ & Recent opened accounts in the last 12 months. (1: Have opened a new account in the last 12 months, 0: No new accounts)  \\ \hline
$X_8$ & Bankruptcies. (1: have declared bankruptcies, 0: Never declared bankruptcies)\\ \hline
$X_9$ & Loan to income ratio. See caption.\\ \hline
$X_{10}$ & Terms. (1: 60 months term, 0: 36 months term)\\\hline
\end{tabular}
\hspace{0.25in}
\vspace{0.05in}
\caption[]{Attributes for the loan issuance data. All variables are Boolean. The outcome $Y$ is a Boolean vector that take the value 1 if the loan fully paid, and 0 otherwise (charged off or marked as default). The loan to income ratio is the ratio $r = \frac{\text{the loan amount}}{\text{annual income}}$, and it is formed as a Boolean function such that $X_9 = \begin{cases} 1, r > \mu \\ 0, r \leq \mu \end{cases}$, where $\mu$ is a threshold ratio that we selected to be the median value of the ratio of all the available dataset, and it was $\mu = 0.2$.}
\label{table:LoanData}
\end{table}

\begin{table}[h]
\centering
\begin{tabular}{ | c | c | c || c || c || c |}
\hline    
$X_3$ & $X_9$ & $X_{10}$ &  Occurrence $P_o$ & $Pr(Y=0 | X)$ & $Pr(Y=0, X)$\\ \hline
0 & 0 & 0 & $38.75\%$ & $24.83\%$ & $9.6\%$\\
0 & 0 & 1 & $5.93\%$ & $33.66\%$ & $1.9\%$\\
0 & 1 & 0 & $23.29\%$ & $34.83\%$ & $8.1\%$\\
0 & 1 & 1 & $15.80\%$ & $39.92\%$ & $6.3\%$\\
1 & 0 & 0 & $4.35\%$ & $31.72\%$ & $1.4\%$\\
1 & 0 & 1 & $0.98\%$ & $38.59\%$ & $0.4\%$\\
1 & 1 & 0 & $5.78\%$ & $41.46\%$ & $2.4\%$\\
1 & 1 & 1 & $5.13\%$ & $46.28\%$ & $2.4\%$\\
\hline
\end{tabular}
\hspace{0.25in}
\vspace{0.05in}
\caption[]{Inferred Boolean relations for the outcome variable $Y$. Each entry in the ``occurrence'' column shows the fraction of observed attribute data $(X_3,X_9,X_{10})$. Given each attribute pattern, the conditional probability that value of outcome $Pr(Y = 0 | X = (x_3,x_9,x_{10}))$, meaning that the probability that the borrower will not fully pay the loan, is shown in the $Pr(Y=0)$ column. The most important quantity that should be consider in the analysis is the joint probability for the pattern occurrence and the outcome $Y = 0$, $Pr(Y = 0 , X = (x_3,x_9,x_{10}))$ is shown in last column of the table.}
\label{table:loanResults}
\end{table}

\section{Discussion and Conclusion}
Although black-box machine learning methods become increasingly more popular due to their relative ease to implement without deep understanding of how they work, in some applications such as quantitative biology where it is essential to uncover causal and relevant factors beyond functional fitting. A prototype problem of such is to learn, from noisy observational data, the structure and function of a Boolean network. The classic widely used REVEAL approach accomplishes this by performing a combinatorial search in the space of Boolean variables, and its performance  relies heavily on having a relatively small network size and small maximum degree, two aspects that are in sharp contrast to typical biological systems that can be large and complex. To overcome these difficulties, here we present BoCSE as a new learning approach based on the optimization of causation entropy applying to Boolean data. This new approach relies on computing entropies of judiciously constructed subsets of variables, and does not require the combinatorial search typically used in other methods. We benchmark  effectiveness of BoCSE on random Boolean networks, and further apply it in several real-world datasets, including in finding the minimal relevant diagnosis signals, quantifying the informative signs of a board game Tic-Tac-Toe, and in determining the causal signatures in loan defaults. In all examples, the BoCSE provides outcomes that is directly interpretable and relevant to the respective application scenarios.

\appendix
\section{Basic concepts from information theory}
In this appendix we review some basic concepts from information theory. These concepts are rooted in information and  theory~\cite{shannon1948,cover2006}, and are heavily utilized in our computational approach for Boolean inference.

The (Shannon) {\it entropy} of a discrete random variable $X$ is given by
\begin{equation}
	H(X) = -\sum_{x}P(x)\log P(x),
\end{equation}
where $P(x)=\mbox{Prob}(X=x)$  and the summation is over the support of $P(x)$, that is, all values of $x$ for which $P(x)>0$. The base of the ``$\log$" function is typically chosen to be 2 so that the unit of entropy becomes ``bit", although other base values can also be used depending on the application.
Entropy is a measure of ``uncertainty" associated with the random variable: generally the larger the entropy is the more difficult it is to ``guess" the outcome of a random sample of the variable.

When two random variables $X$ and $Y$ are considered, we denote their joint distribution by $P(x,y)=\mbox{Prob}(X=x,Y=y)$ and conditional distributions by $P(y|x)=\mbox{Prob}(Y=y|X=x)$ and $P(x|y)=\mbox{Prob}(X=x|Y=y)$, respectively. These functions are used to define the {\it joint entropy} as well as the {\it conditional entropies}, as:
\begin{equation}
\begin{cases}
\mbox{Joint entropy:~}H(X,Y) = -\sum_{x,y}P(x,y)\log P(x,y),\\
\mbox{Conditional entropies:~}\\
~~\mbox{$Y$ given $X$:~}H(Y|X) = -\sum_{x,y}P(x,y)\log P(y|x), \\
~~\mbox{$X$ given $Y$:~}H(X|Y) = -\sum_{x,y}P(x,y)\log P(x|y).\\
\end{cases}
\end{equation}
While the joint entropy $H(X,Y)$ measures the uncertainty associated with the joint variable $(X,Y)$, the conditional entropy $H(Y|X)$ measures the uncertainty of $Y$ given knowledge about $X$ and similar interpretation holds for $H(X|Y)$. In general, $H(Y|X)\leq H(Y)$ and $H(X|Y)\leq H(X)$, with ``$=$" if and only if $X$ and $Y$ are independent.
Interestingly, the reduction of uncertainty as measured by $H(Y)-H(Y|X)$ coincides with $H(X)-H(X|Y)$, leading to a quantity called the 
the {\it mutual information} (MI) between $X$ and $Y$, given by:
\begin{equation}
	I(X;Y) = H(X) - H(X|Y) = H(Y) - H(Y|X).
\end{equation}
Mutual information is symmetric $I(X;Y)=I(Y;X)$, and also nonnegative: $I(X;Y)\geq0$ with $I(X;Y)=0$ if and only if $X$ and $Y$ are independent.

Finally, the {\it conditional mutual information} (CMI) between $X$ and $Y$ given $Z$ is
\begin{equation}
	I(X;Y|Z) = H(X|Z) - H(X|Y,Z),
\end{equation}
which measures the reduction of uncertainty of $X$ given $Z$ due to extra information provided by $Y$.
Conditional mutual information is symmetric with respect to interchanging $X$ and $Y$, and nonnegative, equalling zero if and only if the conditional probabilities $P(x|z)$ and $P(y|z)$ are independent: $P(x|z)P(y|z)=P(x,y|z)$.

\bibliographystyle{siamplain}
\bibliography{cse_refs}

\end{document}